# Iris super-resolution using CNNs: is photo-realism important to iris recognition?

*Eduardo Ribeiro*[1,2] ✉, *Andreas Uhl*[1], *Fernando Alonso-Fernandez*[3]
[1]*Department of Computer Sciences, University of Salzburg, Jakob Haringer Strasse 2 5020, Salzburg, Austria*
[2]*Department of Computer Sciences, Federal University of Tocantins, 109 Norte, Av. NS 15, ALC NO 14, Palmas, Brazil*
[3]*IS-Lab/CAISR, Halmstad University, Box 823, Halmstad SE 301-18, Sweden*
✉ *E-mail: uft.eduardo@uft.edu.br*

**Abstract:** The use of low-resolution images adopting more relaxed acquisition conditions such as mobile phones and surveillance videos is becoming increasingly common in iris recognition nowadays. Concurrently, a great variety of single image super-resolution techniques are emerging, especially with the use of convolutional neural networks (CNNs). The main objective of these methods is to try to recover finer texture details generating more photo-realistic images based on the optimisation of an objective function depending basically on the CNN architecture and training approach. In this work, the authors explore single image super-resolution using CNNs for iris recognition. For this, they test different CNN architectures and use different training databases, validating their approach on a database of 1.872 near infrared iris images and on a mobile phone image database. They also use quality assessment, visual results and recognition experiments to verify if the photo-realism provided by the CNNs which have already proven to be effective for natural images can reflect in a better recognition rate for iris recognition. The results show that using deeper architectures trained with texture databases that provide a balance between edge preservation and the smoothness of the method can lead to good results in the iris recognition process.

## 1 Introduction

The main goal of super resolution (SR) is to produce, from one or more images, an image with a higher resolution (with more pixels) at the same time that produces a more detailed and realistic image being faithful to the low-resolution (LR) image(s). One of the most used examples is bicubic interpolation that, despite producing more pixels and being faithful to the image at LR, does not produce more detailed texture details generating more noise or blur than realism [1].

Several applications, especially in the pattern recognition area, demand, in an ideal environment, images in high resolution (HR) where details and textures from the images may be critical to the final result [2]. With the popularisation of devices built with simpler sensors such as the charged-couple device and the complementary metal oxide semiconductor (CMOS), millions of images have been generated opening a range of possibilities for the most diverse purposes in this area. One of them is biometrics as, e.g. face and iris recognition using mobile phone devices. Biometrics is a very strong and reliable approach for the automatic identification of individuals based on biological phenomena which can be statistically measured. In some practical applications, the lack of pixel resolution in images supplied by less robust sensors (such as mobile phones or surveillance cameras) and the focal length may compromise the performance of recognition systems [3]. In [4], a significant recognition performance degradation is shown when the iris image resolution is reduced.

There are currently two approaches to the SR problem. The first one is based on the use of sub-pixels obtained from several LR images to reach a HR image, also known as reconstruction-based SR [2, 5]. The main disadvantage of this technique is the requirement of multiple images as input to obtain the final image which may make the process unfeasible [6]. The second approach (i.e. also the main focus of this work) called learning-based approach is based on the learning of a model that maps the relation between LR and HR images through a training image database [2]. The advantage of this method is that there is no need for multiple versions of the same image as the input of the system: a single image is required as input. For this reason, this method can also be called as single-image SR approach [7]. This method also can achieve high magnification factors since the model training can be modelled with good performance specially using deep learning approaches.

The use of deep learning, specifically convolutional neural networks (CNNs) to perform the mapping between LR and HR images/patches have been extensively explored in recent years. One of the advantages of using a CNN is that it does not require any handcrafted or engineered feature extractor as those required in previous methods. In addition, the image reconstruction overcomes the performance of traditional methods, particularly in relation to the quality of image textures. However, in the biometrics field, few studies were made exploring this better quality artificially created with respect to the recognition performance.

In this study, we investigate the use of deep learning SR (DLSR) applied to iris recognition. For this, we test different architectures trained from scratch using different databases. The motivation for this is to verify if the proven effectiveness of these methods in relation to the image quality will be reflected in the recognition performance. In addition, through different training databases, we have verified that texture transfer learning can be an alternative to the training of CNNs in practical applications. Specifically, the idea of this contribution is to evaluate if the CNN training with Iris images can specifically learn iris patterns to help in the SR for very small factors where the patterns cannot be identified due to the lack of information and to the image blur. The results show that for very small factors, the databases trained with iris images can achieve better recognition results despite the fact that the present worse quality in the quality assessment algorithm context. This dichotomy is also the main idea of this work contribution and will be discussed in this study.

## 2 Related works

Single-image SR has become the focus of SR discussions in recent years, deriving some surveys about it [8, 9]. Nonetheless, this area has been discussed for decades, beginning with prediction-based methods through filtering approaches (e.g. bilinear and bicubic), which produce smooth textures leading to the study of methods based on edge-preservation [10, 11]. Learning-based (or Hallucination) algorithms using a single image were first introduced in [12] where the mapping between the LR and HR image was learned by a neural network applied to fingerprint images.

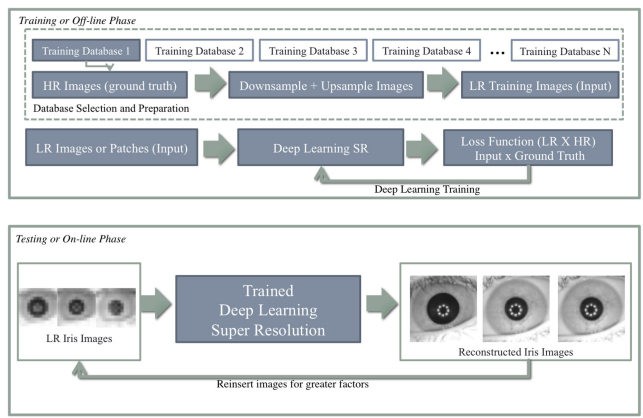

**Fig. 1** *General overview of the training and reconstruction method for the Iris SR using CNNs proposed for this work*

With the popularisation of CNNs, several methods were proposed for obtaining excellent results. Wang *et al.* [13] showed that encoding a sparse representation, particularly designed for SR can make the end-to-end mapping between the LR and HR image through a reduced model size. However, the most famous architecture of this end-to-end mapping is the super-resolution CNN (SRCNN) proposed by Dong *et al.* [14] that used a bicubic interpolation to up-sample the input LR image using a trained three-layer deep fully CNN to reconstruct the HR image acting as a denoising tool. The most common concern of the work that followed was to find an architecture that minimises the mean squared error (MSE) between the reconstructed HR image and the ground truth. Besides that, also reflecting the maximisation of the peak signal-to-noise ratio (PSNR), one of the most used metrics is to evaluate the quality of the result in comparison with the proposed methods [15].

In [16], a deeper CNN architecture is presented inspired by VGG-net used for ImageNet classification [17] also called very deep CNN (VDCNN). That work demonstrates that the use of the cascading of small filters many times in a deep network structure and the use of residual-learning can affect the accuracy of the SR method.

In [15], a SR generative adversarial network (SRGAN) was proposed to try and recover finer texture details from LR images inferring photo-realistic natural images through a novel perceptual loss function using high-level maps from the VGG network. The SRCNN, VDCNN and SRGAN architectures will be used in this work and will be detailed in the following sections.

Research on SR in biometrics (especially for Iris recognition) has been increasing in the last few years specially using reconstruction-based methods. For example, Kien *et al.* [3] use the feature domain to super-resolve LR images relying only on the features incorporating domain-specific information for iris models to constrain the estimation. In [18], Nguyen *et al.* introduce a signal-level fusion to integrate quality scores to the reconstruction-based SR process performing a quality weighted SR for a LR video sequence of a less constrained iris at a distance or on the move obtaining good results. However, in this case, as in [19] that perform the best frame selection, many LR images are required to reconstruct the HR image which is one of the disadvantages of this kind of reconstruction-based methods.

In [20], an iris recognition algorithm based on the principal component analysis (PCA) is presented by constructing coarse iris images with PCA coefficients and enhancing them using SR. In [21], a reconstruction-based SR is proposed for iris SR from LR video frames using an auto-regressive signature model between consecutive LR images to fill the sub pixels in the constructed image. In [22], two SR approaches are tested for iris recognition, one based on the PCA eigen-transformation and other based on locality-constrained iterative neighbour embedding (LINE) of local image patches. Both methods use coupled dictionaries to learn the mapping between LR and HR images in a very LR simulation using infrared iris images obtaining good results for very small images.

Despite the vast literature in the SR area and the great interest in the use of deep-learning in biometrics, the application of DLSR in iris recognition is still an unexplored field, mainly because approaches generally focus on general and/or natural scenes to produce the overall visual enhancement and produce better quality images regarding to photo-realism, while iris recognition focuses on the best recognition performance itself [1, 23]. In [24], three multilayer perceptrons (MLPs) are used to perform single image SR for iris recognition. The method is based on merging the bilinear interpolation approach with the output pixel values from the trained multiple MLPs considering the edge direction of the iris patterns. Recently, Zhang *et al.* [25] use the classic SRCNN and SR forest to perform SR in mobile iris recognition systems. The algorithms are applied to the segmented and normalised iris images and the results show a limited effectiveness of the SR method for the iris recognition accuracy. Different from the methods presented in the DLSR literature, in this work, we explore if the architectures and database used in training can have an influence on the quality results, and consequently on the recognition performance.

In our previous works [26, 27], we demonstrated that basic deep learning techniques for super-resolution such as stacked autoencoders and the classic SRCNN can be successfully applied to Iris SR. In that case, we used the CASIA Interval database as a target database focusing more on the recognition process. In this work, we focus on the relation between the quality and the performance of the recognition and the SR is performed on the original image without any segmentation. We also use a new iris database as a target database that simulates a real world situation where the images are acquired using mobile phones. Additionally, we test a new application that is the use of SRGANs to verify if the good performance of this method for natural images in terms of photo-realism is also valid and useful for iris images in the iris recognition context.

## 3 Reconstruction of LR iris images through CNNs

Typically, in a deep learning system, the main question is to find a good training database that can provide relevant information to the desired application. In the case of SR, it is necessary to achieve, during the proposed method training (also called the off-line phase), a mapping between a high-resolution (HR) image with high-frequency information and a LR image with low-frequency information. Fig. 1 shows this phase, in which a training database is chosen and the images are prepared for the deep learning SR method training.

In the training phase, the only pre-processing required is, given an image in HR X, that the image needs to be downscaled to one or more factors followed by an up-scaling using bicubic interpolation to the same size as the original image X. This image, although it has the same size as X is called 'LR' image and is denoted as the LR image Y. The purpose of DLSR training is, after feeding the network with an LR image or patch Y as input, try to obtain a result F(Y) (the reconstructed image) as similar as possible to the HR image or patch X, in this case, the ground truth. The weight adjustment of the method will depend on both the chosen architecture and the loss function that will be better explained in the following sections.

After training, the deep learning method is applied to a LR database for the proposed application which is, in the case of this work, an iris database also called target database. If so, the deep learning process is a pre-processing step before the iris recognition, in which the LR image is introduced as an input to the network that will produce the reconstructed image in the HR to be used in the recognition process as is shown in Fig. 1 (on-line phase) that will be reconstructed based on the factor training.

In deep learning, the preparation of individual machines for all possible scenarios to deal with different scales, poses, illumination, and textures is still a challenge. In this work, we test the main SR architectures, using different databases for the training, to evaluate questions such as: if the similarity of the training database with the target database can aid in the process of SR or if the use of a distinct part of the target database itself (obtained during the enrolment of the individuals) can be used and if this knowledge can be transferred in a practical application.

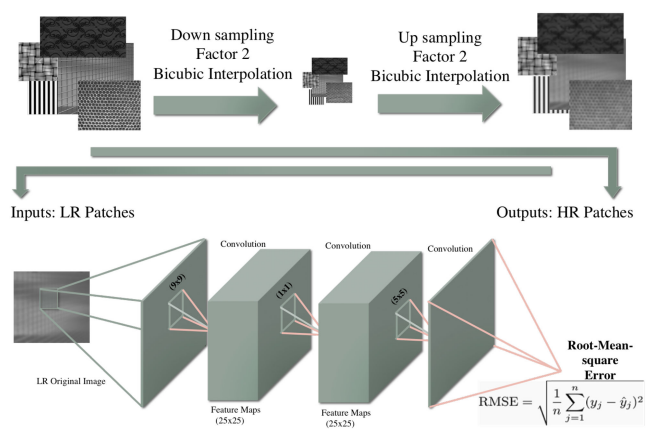

**Fig. 2** *Illustration of the CNN architecture for Iris SR (SRCNN)*

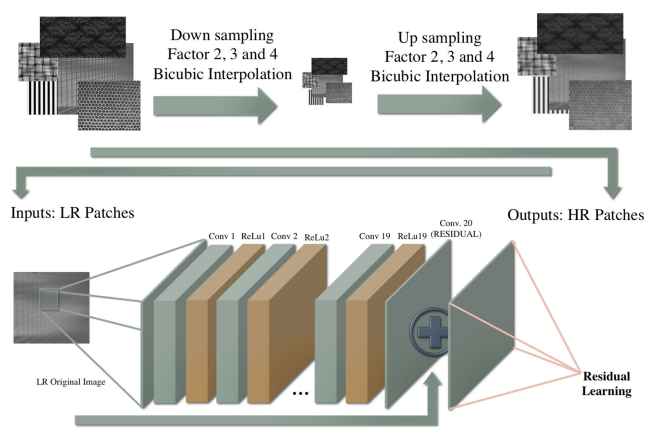

**Fig. 3** *Illustration of the VDCNN architecture for Iris SR*

## 4 CNN architectures

CNNs are considered the evolution of traditional neural networks, however, they share the same essence: a map of neurons with learnable weights, biases, activation functions and loss functions. The main impulse that contributed to the CNN popularity was the capability of treating 3D volumes of neurons (width, height, and depth). Generally, the input of a CNN is formed by this 3D volume of size $m \times m \times d$, e.g. an image, where $m \times m$ is the dimension of the image and $d$ is the number of channels. The architecture of a CNN is defined depending on the application and generally is constructed stacking those layers using three main types: convolutional layers, pooling layers and fully connected layers (exactly as seen in the traditional Neural Networks). A convolutional layer is formed by a series of $k$ learnable filters with size $(n \times n \times d)$ where $(n \leq m)$. These filters (also known as kernels) are convolved in the input volume resulting in the so-called activation maps or feature maps. As classic neural networks, the convolution layer outputs are subjected to an activation function, e.g. the ReLU rectifier function $f(x) = \max(0, x)$, where $x$ is the neuron input.

In this work, we use three different deep learning approaches for SR: SRCNN, VDCNN, and SRGAN. For each of these approaches, architectures and methodologies used to conduct the image reconstruction are explained in the next subsections.

### 4.1 SR convolutional neural network (SRCNN)

One of the first CNN architectures in SR presented was the SRCNN [14] (see Fig. 2). This classical approach consists of three layers representing: the patch extraction, a non-linear mapping, and the reconstruction step. As a pre-processing step, patches of size $33 \times 33$ (also called HR patches) are extracted from the training images, then, as mentioned in the previous section, the patches are downscaled for the factor two and up-scaled for the original size using bicubic interpolation. These also called LR patches are used as the input for the CNN in the training phase.

The SRCNN architecture is specified as follows: the first layer (patch extraction) consists of 64 filters of size $9 \times 9 \times 1$ with stride one and padding zero, the second layer (non-linear mapping) has 32 filters of size $1 \times 1 \times 64$ with stride one and padding zero, and the last layer (reconstruction) has one filter of size $5 \times 5 \times 32$ with stride one and padding zero. The loss-function used in the CNN training is the MSE between the output (reconstructed patch) and the ground truth (HR patch). Loss minimisation is done using stochastic gradient descent, the MatConvNet framework [28] is used for implementation.

### 4.2 Super-resolution very deep convolutional neural network (VDCNN)

This architecture proposed in [16] relies on the use of a deeper CNN inspired by the VGG-net used for ImageNet classification (see Fig. 3). In the training phase, a pre-processing step is done by extracting HR patches and downscaling them by factors of two, three and four, re-up-scaling them to the same size as the HR patches serving as the input for the CNN (LR patches).

The VDCNN architecture is composed of 20 layers with the same parameterisation (except for the first and last layers): 64 filters of size $3 \times 3 \times 64$. In this approach, a residual learning introduced by He *et al.* [29] is used to avoid the degradation problem when the network depth increases. In the residual learning, the degradation needs to be minimised in order that the deeper model should produce no higher training error than its shallower counterpart. Instead of using the traditional MSE error between the original image and the reconstructed one hoping that the added layers can fit in the mapping, this method allows these layers to fit the residual mapping that is easier and faster to optimise. The loss function used in the training is the MSE between the residual input error (the difference between the reconstructed patch and the HR patch) and the residual ground truth that, in this case, is the difference between LR and HR patch.

This residual-learning boosts the convergence of the CNN training and, consequently, its performance. The loss minimisation is done also based on the gradient descent with back-propagation [30] using the MatConvNet framework [28].

### 4.3 SR generative adversarial network (SRGAN)

This architecture proposed in [15] relies on two different CNNs with a new scheme of objective functions in an attempt to recover finer texture details from very LR images. While the generator architecture is responsible for generating the HR reconstructed image from the LR one, the discriminator architecture is trained to differentiate the reconstructed image from the original photo-realistic one.

As it can be seen in Fig. 4, the generator network is basically a series of residual blocks with identical layouts: two convolutional layers of size with $3 \times 3 \times 64$ followed by a batch-normalisation and parametric-RELU layers as the activation function. The main objective of this approach is the training of a generative model with the goal of 'fooling' a proposed adversary (differentiable discriminator) trained to discriminate super-resolved images from real images [31].

The discriminator network is also based on the VGG network and contains eight convolutional layers with filters of size $3 \times 3 \times T$, where $T$ is increased by a factor 2 through the layers from 64 to 512 filters as in the VGG network. The discriminator is trained in order to maximise the correct label for the training images and the samples reconstructed by the generative model by a method called perceptual loss function [31]. This loss function uses the output of both CNNs (content loss and adversarial loss) trying

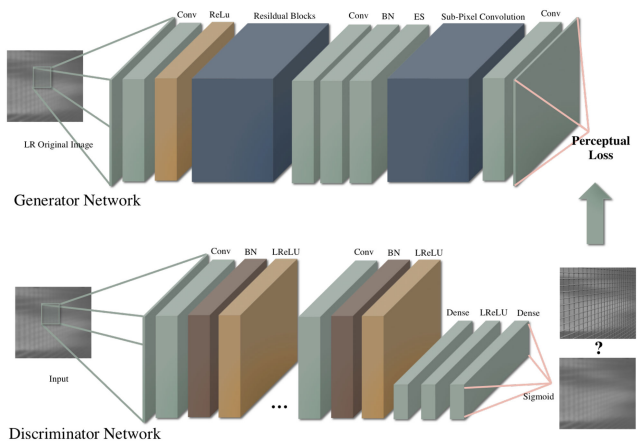

**Fig. 4** *Illustration of the GAN architecture for Iris SR (SRGAN)*

to assess a solution with respect to perceptually relevant characteristics.

For the training, the images were cropped to a size of 96 × 96 pixels and down-sampled by factor 4 (LR input) as a pre-processing step.

## 5 Databases

### 5.1 Target/test database

In this work, we use as the target database one of the most widely used databases in biometrics experiments: the CASIA Interval V3 database. This database contains 2655 near infrared images of size 280 × 320 from 249 subjects captured with a self-developed close-up camera, resulting in 396 different eyes which will be considered as different subjects for this work. For the experiments, all the images from this database are interpolated using bicubic interpolation in order to have the same sclera radius followed by a cropping around the pupil in a square region of size 231 × 231. When the images do not fit in this cropping (e.g. if the iris is close to a margin), they are discarded. With this pre-processing step, 1872 images from 249 users remained in the database.

In the experiments we explore texture transfer learning among different databases, which means that the CNN is pre-trained with a different database (texture, natural or iris database), then it is used to perform the SR in the target image database. For this part, we divide the target database into two: one with the first three images of each user (representing the enrolment images in a real world situation), and the other with the remaining images from each user (representing the authentication database). The registration database is one of the iris training databases among the other texture and natural databases that are explained in the next section.

### 5.2 Training databases

As mentioned in the previous section, for CNN training we use ten different databases of different nature to test transfer learning and its impact on the recognition process. The databases include four texture datasets, two natural image datasets and four iris datasets (from the public IRISSEG-EP [32] dataset) detailed as follows.

*5.2.1 Texture databases:* The Amsterdam Library of Textures (ALOT) with 27,500 rough texture images of size 384 × 256 was divided into 250 classes [33]. The describable texture dataset (DTD) with 5640 images of sizes range between 300 × 300 and 640 × 640 categorised into 47 classes [34]. The Flickr material database (FMD) contains 1000 images of size 512 × 384 divided into 10 categories [35]. The textures under varying illumination, pose and scale (KTH-TIPS) database with 10 different materials containing 81 cropped images of size 200 × 200 in each class [36].

*5.2.2 Natural image databases:* The CALTECH101 Database is a natural image dataset with a list of objects belonging to 101 categories [37]. The COREL1000 database is a natural image database containing 1000 colour photographs showing natural scenes of ten different categories [38].

*5.2.3 Iris databases:* The IIT Delhi Iris Database (IITD) is an Iris Database consisting of data acquired in a real environment resulting in 2240 images of size 230 × 240 from a digital CMOS near-infrared camera. The CASIA-Iris-Lamp (CASIAIL) is an Iris database collected using a hand-held iris sensor and containing 16,212 images of size 320 × 280 with nonlinear deformation due to the variations of visible illumination. The UBIRIS v2 Iris database is a database containing 2250 images of size 400 × 300 captured on non-constrained conditions (at-a-distance, on-the-move and on the visible wavelength), attempting to simulate more realistic noise factors. The NOTREDAME Iris Database is a collection of close-up near-infrared Iris images containing 837 images of size 640 × 480 with off-angle, blur, interlacing, and occlusion factors.

## 6 Experimental setup

For the experiments, we test different down-sampling factors for the target database. For example, if the original image has a size of 231 × 231 and is down-sampled for factor 4, this will correspond to 16× reduction in image pixels in a new image of size 57 × 57. Regardless of the chosen factor, for comparison criteria, all images are reconstructed by the CNNs until they reach the original size.

All methods evaluation and comparison in all stages of this work are based on the quality evaluation of the images as well as on the accuracy of the iris recognition. The qualitative assessment data will be given by two measures: The peak signal to noise ratio (PSNR) and the structural similarity index measure (SSIM) where a high score reflects a high quality using the HR image as the reference image.

For the recognition experiments, we use one iris segmentation algorithm and two different feature extraction methods from the USIT − University of Salzburg Iris Toolkit v2 for iris recognition [39]. In the segmentation process, the iris is segmented and wrapped to a normalised rectangle of size 64 × 512 through the weighted adaptive Hough and ellipsopolar transform (WAHET). The first feature extraction is based on a complex Gabor filter bank with eight different filter sizes and wavelengths Complex Gabor Filterbanks (CG) while the second method is a classical wavelet-based feature extraction with a selection of spatial wavelets Quadratic Spline Wavelet (QSW). In both cases, the bit-code vectors are compared using the normalised Hamming distance [39]. Using the target database (CASIAIrisV3- Interval) with 249 users containing at least five or more images per user, we obtain 5087 genuine and 1,746,169 impostors scores.

## 7 Experimental results

### 7.1 Texture transfer learning comparison

In this section, we explore the use of texture transfer learning as an alternative to the training of CNNs in practical applications. For this, we chose to use the most basic architecture (SRCNN) trained with ten different databases, including texture databases (ALOT, DTD, FMD and KTH-TIPS), natural image databases (CALTECH 101, COREL1000 and IITD) and Iris databases (CASIAIL, UBIRIS and NOTREDAME) applying it to the target database (CASIA Interval). In all frameworks, for a fair comparison between the databases, a subset of 150,000 patches are extracted from each database to pre-train each CNN from scratch, when the CNN weights are initialised randomly.

We also compare the results with the use of two basic interpolation methods: bilinear and bicubic interpolation. Besides that, we train the CNN with the remaining images from the target database after the splitting (as explained in the target database section) to compare if images from the same individual can be beneficial to the CNN training.

In Table 1, we present the quality assessment results for the transfer learning in these databases for different factors: 2, 4, 8 and 16. In all the cases, the images were downscaled for these factors and reconstructed using the CNNs trained with the chosen

72

databases shown in each column in the table. It can be noticed that the quality of the reconstructed images is more similar to the HR images than the interpolated ones for all factors, including larger down-sample factors such as factors 8 and 16, demonstrating the flexibility of deep-learning when image resolution decreases. It also can be seen that the group with the best performance is the texture database group showing that the texture patterns can provide a better generalisation for iris texture reconstruction (surprisingly, even better than most iris datasets). On the other hand, the results from using the CASIA Interval dataset (with different images from the same individual for training and testing) also present a good quality compared with the other databases.

In Table 2, we present the results for iris recognition in order to be able to relate the photo-realism of the reconstructed images with the iris recognition performance. It can be seen that the best results were diversified among the methods and training databases also showing a divergence from the best results presented by the quality performance. The only database with the best result, both for the quality assessment results and the recognition accuracy is the DTD database for factor 2 (115 × 115) with 6.657% of Equal Error Rate (EER). Another interesting point to notice is that, for factors 2 and 4, almost all reconstruction methods surpass the results using original images (the results for the original images are in Table 2 caption) including the bicubic interpolation which is, for small factors, better than all the CNN results.

Using the enrolment images from the same target database (CASIA Interval) does not lead to good recognition performance, which means that the CNN poorly memorise the actual patterns of the users focusing more in general patterns, mainly because of the depth of the network that does not allow a high feature discrimination.

### 7.2 Architectures comparison

To compare the three different CNN approaches, we take into consideration two databases from the transfer learning experiments: the DTD database that presented, in general, a good performance both for quality and recognition measures and the CASIA Interval database that uses the same database divided into training (simulating registration/enrolment images already stored in the system in a real situation) and testing (simulating the verification images) to see how the other CNN behave with the same iris patterns presented in the training.

In this experiment, we test all the architectures explained in Section 6 analysing the performance using quality assessment algorithms as well as the recognition performance, which are presented in Table 3 and the visual results presented in the example in Fig. 5. It can be seen that there is not a simple best approach for all the factors showing that there is not a 'universal approach and general training database' to be used that can lead to the best results for quality and recognition process in all factors.

It is interesting to notice that, in case of down-sampling the images by factor 4 and reconstructing them, the results are better than using the original images (the results for the original images are in Table 3 caption). This means that, in terms of recognition, it is better to downscale the original image (i.e. apply a blur filter) and apply the deep-learning methods before recognition to perform a kind of denoising process in order to achieve better results for the recognition algorithms.

It also can be noticed that, for larger factors, the best approach using the quality assessment algorithms as a comparison measure is the VDCNN using the DTD database as training. However, for the recognition algorithms, the best results were divided between the approaches. For factor 8, the SRGAN architecture presents a great result compared with the other approaches for the CASIA Interval database showing that deeper layers are capable of extracting more specific texture patterns from the users showing much better and consistent performance with this CNN. It is also worthy to notice that for very small images (specially for factor 16) the difference between the methods, in this case, is not debatable since the accuracy above 30% is unacceptable for a recognition system.

We also calculate the density distributions of the similarity scores in a genuine-impostor comparison for the target dataset (CASIA Interval V3 database) and for different architectures (Very Deep Convolutional Network (VDCNN) and SRGAN) trained with the DTD database that presented the best recognition results. We analyse the distribution among different scaling factors using the QSW features that are shown in Figs. 6 and 7. It can be seen in both figures (subfigures (a), (b) and (c)) that the distribution curves overlap only a little bit, making it easy to define the decision threshold clearly differentiating authorised and unauthorised users. This kind of distribution shows that there is no big degradation in the patterns until factor 4, however, using the 8 and 16 it can be seen that the distance between the match and non-match distributions decreases, making it difficult to make the decision and consequently decreasing the recognition performance. Comparing the two architectures, it can be seen that the probability density for small factors is almost the same, however, when the image size decreases the VDCNN architecture seems to present less shift to the right in the genuine scores showing that this CNN can reconstruct images with more differentiable patterns.

From this figure, the intersection between the genuine and imposter distributions for the CNN methods for factor 2 is smaller than the one when using the original images. This result shows that the approach for small factors acts as an image de-noising, producing fewer errors in the recognition performance. This phenomenon can be explained by the fact that the reconstructed images of different users tend to be more 'similar' (under the features used here) as they pass through the same down-sampling and up-sampling procedure.

The example image in Fig. 5 allows comparing the photo-realism presented by the methods in each factor. It can be noticed that the SRGAN approach tries to maximise the edge preservation generating a more consistent photo-realism when the factor increases. However, this leads to too many artefacts that can lead to poor results in the recognition process. As it can see by the red-squared images, the recognition performance is better when there is a balance between the texture, edge preservation, and photo-realism of the iris.

### 7.3 Methods comparison using mobile phone images databases

In this section, we explore the use of CNNs in a real world situation where the images are captured from mobile devices, comparing our results with another method found in the literature for iris SR (PCA-SR). For a complete comparison, in this case, we use three quality assessment algorithms and two different recognition approaches (also used in the iris SR literature) that will be explained next.

For this experiment, we chose the visible spectrum smart-phone Iris (VSSIRIS [40]) database with images captured using two different mobile phone devices: Apple iPhone 5S (3264 × 2448 pixels) and Nokia Lumia 1020 (3072 × 1728 pixels). For each device, five images of the two eyes from 28 subjects were captured totalling 280 images per device or 560 in total. Fig. 8 shows some example images from each device. As a pre-processing step, all the images are resized to have the same sclera radius followed by a cropping around the pupil in a square region of size 319 × 319 pixels. The down-sampling factor used for this experiment is factor 22 following the previous studies in [41, 42] to generate a very small iris region (13 × 13 pixels) for the real-world LR simulation.

We compare the CNN (trained with the DTD database) results with a method used in [22, 42] for iris-SR called PCA hallucination of local patches based on the algorithm for face images of [43] where a PCA eigen-transformation is conducted on the set of LR basis patches to use the weights provided by the projection of the eigen-patches to reconstruct the images.

The PCA algorithm uses a training database where each image $X$ is subdivided in LR overlapped patches $X = x_1, x_2, \ldots, x_N$ forming the so-called LR basis patches $L_i$. Two super sets of basis patches $H_i$ and $L_i$ are computed for each patch $x_i$ from this database. Given an input patch $x_i$, it will be projected onto the eigen-patches of $L_i$, obtaining the optimal reconstruction weights $c_i = \{c_i^1, c_i^2, \ldots, c_i^M, \}$ of $x_i$ w.r.t $L_i$. After obtaining the optimal reconstruction weights $c_i$, the HR patches are super-resolved using the $y_i = H_i c_i^t$ where $y_i$ are the overlapped reconstructed patches to form the preliminary reconstructed HR image $Y$. This linear combination of collocated HR patches employing all available collocated patches of the training database with their own optimal reconstruction coefficients, allow the preservation of local image information making a good reconstruction. The problem with this method is that it requires the use of the same database for training and testing making difficult the use of practical and real world applications. Another problem is the assumption that reconstruction weights are

**Table 1** Results of quality assessment algorithms for texture transfer learning comparison with different downscaling factors (average values on the test dataset) using the SRCNN architecture compared with the bilinear and bicubic approach

| LR size (scaling) | | Bilinear | Bicubic | Texture databases | | | | Natural image databases | | | Iris databases | | |
|---|---|---|---|---|---|---|---|---|---|---|---|---|---|
| | | | | ALOT | DTD | FMD | KTH-TIPS | CALTECH 101 | COREL 1000 | IITD | CASIAIL UBIRIS | NOTREDAME | CASIA Interval |
| 115 × 115 (1/2) | PSNR | 0.8855 | 0.8957 | 0.9481 | **0.9595** | 0.9509 | 0.9485 | 0.9492 | 0.9491 | 0.9483 | 0.9422 0.9414 | 0.9495 | 0.9502 |
| | SSIM | 30.77 | 31.07 | 35.17 | **35.87** | 35.82 | 35.79 | 35.85 | 35.34 | 35.43 | 35.12 34.67 | 35.70 | 35.80 |
| 57 × 57 (1/4) | PSNR | 0.7949 | 0.8089 | 0.8243 | **0.8259** | 0.8245 | 0.8232 | 0.8250 | 0.8255 | 0.8214 | 0.8129 0.8131 | 0.8216 | 0.8240 |
| | SSIM | 27.99 | 28.67 | 29.20 | **29.32** | 29.29 | 29.23 | 29.24 | 28.97 | 29.18 | 29.01 28.86 | 29.24 | 29.29 |
| 29 × 29 (1/8) | PSNR | 0.6956 | 0.7061 | 0.7198 | 0.7228 | 0.7157 | 0.7204 | **0.7251** | 0.7236 | 0.7127 | 0.7064 0.7085 | 0.7128 | 0.7174 |
| | SSIM | 24.59 | 25.06 | 25.61 | 25.79 | 25.57 | 25.69 | **25.80** | 25.50 | 25.44 | 25.15 25.12 | 25.44 | 25.54 |
| 15 × 15 (1/16) | PSNR | 0.6120 | 0.6160 | 0.6510 | 0.6544 | 0.6471 | 0.6503 | **0.6557** | 0.6553 | 0.6439 | 0.6406 0.6430 | 0.6447 | 0.6494 |
| | SSIM | 20.78 | 20.93 | 23.09 | **23.23** | 23.07 | 23.04 | 23.21 | 23.05 | 23.01 | 22.67 22.69 | 22.97 | 22.95 |

**Table 2** Verification results (EER) for texture transfer learning comparison for different downscaling factors using the SRCNN architecture comparing to the Bilinear and Bicubic approach. The accuracy result for the original database with no scaling is 6.65% for WAHET + CG and 3.81% for WAHET + QSW

| LR size (scaling) | | Bilinear | Bicubic | Texture databases | | | | Natural image database | | | Iris databases | | |
|---|---|---|---|---|---|---|---|---|---|---|---|---|---|
| | | | | ALOT | DTD | FMD | KTH-TIPS | CALTECH 101 | COREL 1000 | IITD | CASIAIL UBIRIS | NOTREDAME | CASIA interval |
| 115 × 115 (1/2) | WAHET + CG | 6.32 | 6.39 | 6.50 | **6.07** | 6.66 | 7.16 | 6.74 | 6.39 | 6.68 | 6.61 6.37 | 6.64 | 6.83 |
| | WAHET + QSW | **3.26** | 3.58 | 3.58 | 3.32 | 3.81 | 4.28 | 4.02 | 3.53 | 3.89 | 3.92 3.42 | 4.02 | 3.84 |
| 57 × 57 (1/4) | WAHET + CG | 9.36 | **5.81** | 7.19 | 6.67 | 6.88 | 6.22 | 6.83 | 6.51 | 7.90 | 7.84 8.41 | 7.59 | 6.66 |
| | WAHET + QSW | 6.10 | **2.65** | 4.58 | 3.78 | 4.09 | 3.62 | 3.95 | 3.74 | 5.11 | 5.22 5.75 | 4.66 | 3.93 |
| 29 × 29 (1/8) | WAHET + CG | 36.11 | 42.22 | 32.97 | 32.19 | 36.86 | **22.41** | 32.88 | 33.81 | 38.19 | 39.88 39.75 | 39.15 | 33.89 |
| | WAHET + QSW | 33.60 | 42.34 | 30.62 | 31.13 | 34.89 | **21.75** | 32.10 | 33.26 | 36.50 | 38.53 37.33 | 37.04 | 30.65 |
| 15 × 15 (1/16) | WAHET + CG | 31.66 | 32.96 | 33.95 | 33.10 | 33.03 | 33.96 | 33.02 | 34.68 | 32.73 | **28.52** 29.62 | 31.50 | 31.57 |
| | WAHET + QSW | 30.68 | 32.18 | 32.57 | 32.06 | 31.60 | 33.06 | 31.66 | 33.18 | 31.84 | **27.60** 28.02 | 31.25 | 30.17 |

**Table 3** Quality assessment (PSNR and SSIM) and verification results (EER, WAHET + CG and WAHET + QSW) for different databases training and different downscaling factors using different architectures. The accuracy result for the original database with no scaling is 6.65% for WAHET + CG and 3.81% for WAHET + QSW

| LR size (scaling) | | Bilinear | Bicubic | CASIA interval | | | DTD | | |
|---|---|---|---|---|---|---|---|---|---|
| | | | | SRCNN | VDCNN | SRGAN | SRCNN | VDCNN | SRGAN |
| 115 × 115 (1/2) | PSNR | 0.8855 | 0.8957 | 0.9502 | 0.9555 | 0.9075 | **0.9595** | 0.9540 | 0.8937 |
| | SSIM | 30.77 | 31.07 | 35.80 | **36.90** | 23.46 | 35.87 | 36.56 | 21.68 |
| | WAHET + CG | 6.32 | 6.39 | 6.83 | 6.63 | 6.70 | **6.07** | 6.32 | 7.71 |
| | WAHET + QSW | **3.26** | 3.58 | 3.84 | 3.78 | 4.27 | 3.32 | 3.53 | 3.94 |
| 57 × 57 (1/4) | PSNR | 0.7949 | 0.8089 | 0.8240 | 0.8347 | 0.7914 | 0.8259 | **0.8348** | 77.47 |
| | SSIM | 27.99 | 28.67 | 29.29 | 29.60 | 24.10 | 29.32 | **29.65** | 22.15 |
| | WAHET + CG | 9.36 | **5.81** | 6.66 | 6.51 | 6.98 | 6.67 | 6.69 | 8.57 |
| | WAHET + QSW | 6.10 | **2.65** | 3.93 | 3.26 | 4.06 | 3.78 | 3.41 | 4.00 |
| 29 × 29 (1/8) | PSNR | 0.6956 | 0.7061 | 0.7174 | 0.7332 | 0.6333 | 0.7228 | **0.7374** | 0.6488 |
| | SSIM | 24.59 | 25.06 | 25.54 | 26.04 | 22.08 | 25.79 | **26.21** | 21.00 |
| | WAHET + CG | 36.11 | 42.22 | 33.89 | 17.88 | **13.58** | 32.19 | 19.07 | 21.09 |
| | WAHET + QSW | 33.60 | 42.34 | 30.65 | 16.72 | **13.38** | 31.13 | 17.07 | 19.50 |
| 15 × 15 (1/16) | PSNR | 0.6120 | 0.6160 | 0.6494 | 0.6563 | 0.5568 | 0.6544 | **0.6633** | 60.79 |
| | SSIM | 20.78 | 20.93 | 22.95 | 23.30 | 20.66 | 23.23 | **23.57** | 20.83 |
| | WAHET + CG | 31.66 | 32.96 | 31.57 | 33.87 | 38.32 | 33.10 | 33.85 | 34.46 |
| | WAHET + QSW | 30.68 | 32.18 | 30.17 | 32.03 | 38.41 | 32.06 | 31.76 | 35.92 |

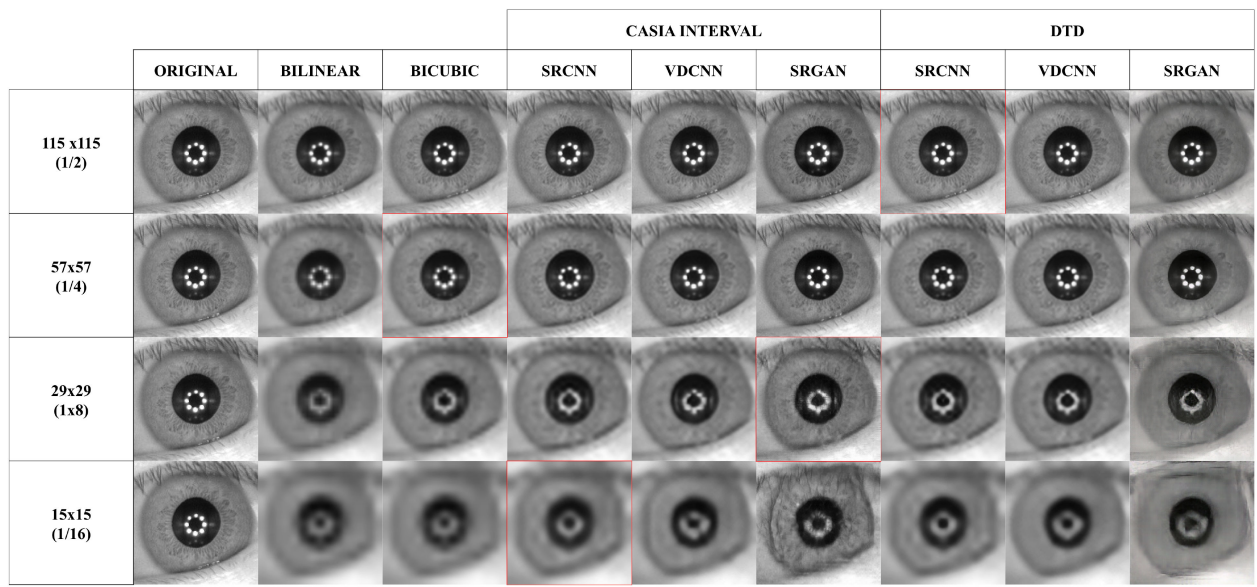

**Fig. 5** *Resulting images for different sampling factors in different approaches. The original image is replicated in all rows in the column. The red-squared images represent the approach that has the best recognition performance for the factor*

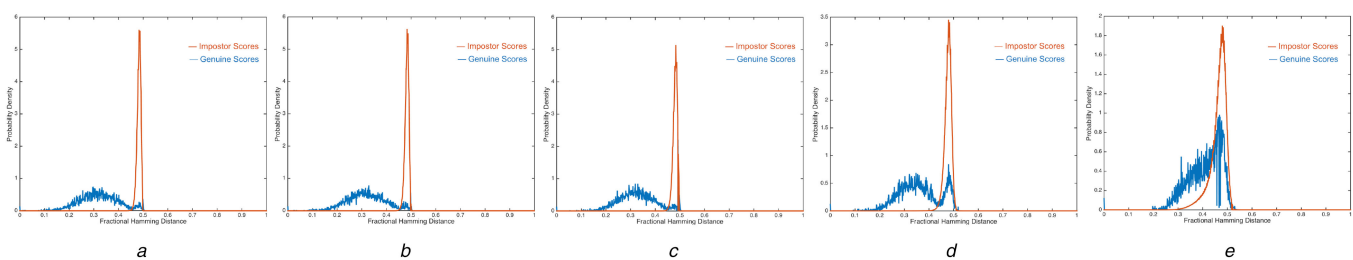

**Fig. 6** *Density distribution histogram of similarity scores for QSW features for the VDCNN trained with the textural DTD database. (a) Original target database, (b)–(e) reconstructed databases using different scales*
*(a)* Original, *(b)* VDCNN 115 × 115 (1/2), *(c)* VDCNN 57 × 57 (1/4), *(d)* VDCNN 29 × 29 (1/8), *(e)* VDCNN 15 × 15 (1/16)

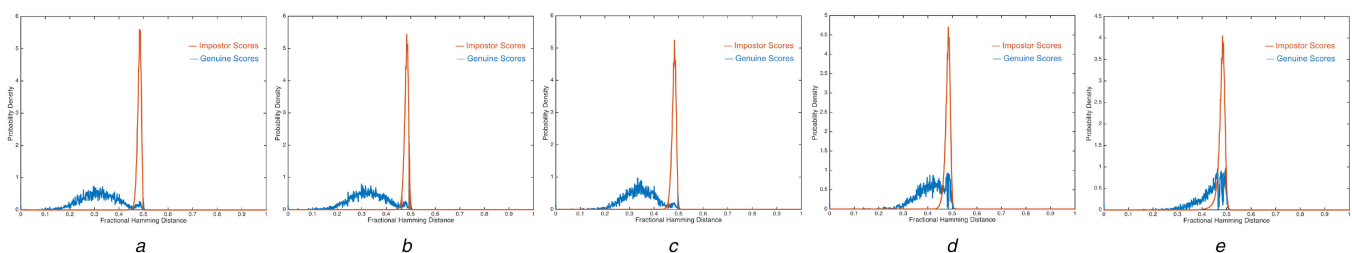

**Fig. 7** *Density distribution histogram of similarity scores for QSW features for the SRGAN trained with the textural DTD database. (a) Original target database, (b)–(e) reconstructed databases using different scales*
*(a)* Original, *(b)* SRGAN 115 × 115 (1/2), *(c)* SRGAN 57 × 57 (1/4), *(d)* SRGAN 29 × 29 (1/8), *(e)* SRGAN 15 × 15 (1/16)

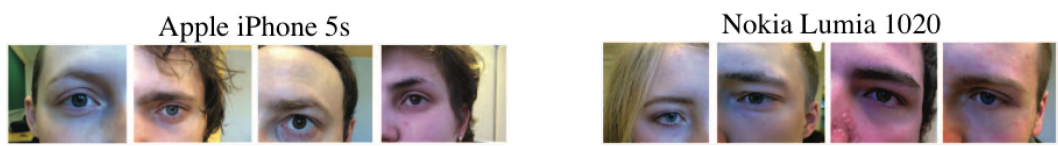

**Fig. 8** *Sample images from VSIRIS database [40]*

the same in the LR and HR manifolds which can be a problem for large databases with millions of different patterns which can generate a distortion by the one-to-many relationship between LR and HR patches.

Besides the two quality assessment measures (PSNR and SSIM) we also compute the feature similarity index for image quality assessment (FSIM) from [44] that extracts low-level features including the significance of local structures and the image gradient magnitude. Table 4 presents the results using these quality assessment metrics comparing the CNNs trained with the DTD database with the bicubic, bilinear and PCA approaches for the full image and for the iris region.

It can be seen that using the traditional measures (PSNR and SSIM) the approach that presented the best results was the PCA reconstruction. However, analysing the images from Fig. 9 it can be noticed that, in fact, the images that present more photo-realism are the images from the SRGAN CNN reconstruction in both devices. The PCA approach, although exhibiting the best results for PSNR and SSIM, visually looks most artificially among all the methods, generating big squares of pixels due to the eigen-patch reconstruction nature.

However, using FSIM that is based on modelling the human visual system, the best results are obtained by the SRGAN CNN. Using this metric and analysing the images from Fig. 9 it can be noticed that, in fact, the images that present most photo-realism are the images from the SRGAN CNN reconstruction for both devices.

**Table 4** Quality assessment (PSNR, SSIM) results for different methods employed in the VSSIRIS database

| LR size (scaling) | | Full image | | | | | Iris region | | | | |
|---|---|---|---|---|---|---|---|---|---|---|---|
| | | Bilinear | Bicubic | PCA | VDCNN | SRGAN | Bilinear | Bicubic | PCA | VDCNN | SRGAN |
| 13 × 13 (1/22) | PSNR | 24.44 | 24.97 | **26.00** | 25.26 | 24.59 | 24.35 | 24.89 | **25.45** | 24.89 | 18.08 |
| | SSIM | 0.7200 | 0.7200 | **0.7300** | 0.7256 | 0.5862 | 0.6200 | 0.6400 | **0.6700** | 0.6476 | 0.5395 |

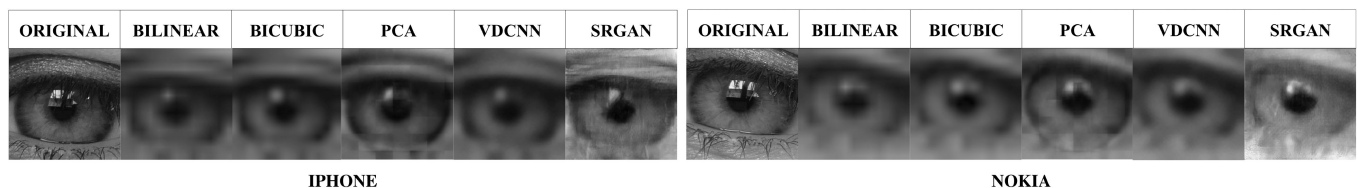

**Fig. 9** *Resulting images for different sampling factors in different approaches using the VSIRIS database*

**Table 5** Verification results (EER) for different methods employed. The accuracy result for the original database with no scaling is 38.36% (WAHET + QSW) and 39.89% (WAHET + CG) for iPhone images, 31.60% (WAHET + QSW) and 36.75% (WAHET + CG) for Nokia images, 0.33% (SIFT) for iPhone images and 0.68% (SIFT) for Nokia images

| LR size (scaling) | | iPhone | | | | | Nokia | | | | |
|---|---|---|---|---|---|---|---|---|---|---|---|
| | | Bilinear | Bicubic | PCA | VDCNN | SRGAN | Bilinear | Bicubic | PCA | VDCNN | SRGAN |
| 13 × 13 (1/22) | WAHET + QSW | 32.64 | 33.16 | 33.80 | **31.17** | 39.29 | 30.78 | 30.81 | 32.40 | **28.11** | 39.09 |
| | WAHET + CG | 35.99 | 35.93 | 35.55 | **32.23** | 42.74 | 31.13 | 31.18 | 36.08 | **27.54** | 41.93 |
| | SIFT | 23.54 | 22.80 | **9.30** | 24.28 | 12.00 | 26.50 | 29.80 | **11.13** | 26.61 | 14.09 |

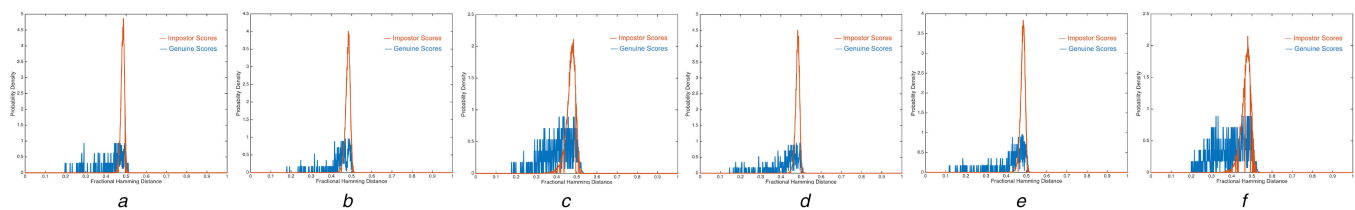

**Fig. 10** *Density distribution histogram of similarity scores for QSW features for the VDCNN and SRGAN trained with the textural DTD database for the scale 1/22 (13 × 13 pixels). (a, d) Original target databases. (b, c, e, f) reconstructed databases*
*(a)* iPhone original, *(b)* iPhone SRGAN, *(c)* iPhone VDCNN, *(d)* Nokia original, *(e)* Nokia SRGAN, *(f)* Nokia VDCNN

For the verification experiments, we use the same recognition algorithms from the last experiments to evaluate the performance of the reconstruction methods. Table 5 presents the EER for these methods with the experiments performed separately for each smartphone with 560 genuine and 38,500 impostor scores per device. We also use a new comparator for this experiment called Scale Invariant Feature Transform (SIFT) comparator [45] in which SIFT feature points are extracted from the iris and compared based on the texture information around the points [46]. This is motivated by the factor that this feature extractor does not need any segmentation stage in the process which would be good for images with low quality. Nonetheless, this feature extraction algorithm was not used in the experiments in Sections 7.1 and 7.2 because it does not provide any additional information for discussion, unlike in the current experiment where the results are different compared with the other feature extraction methods.

It can be noticed that for the WAHET + QSW and the WAHET + CG features, the best results for recognition are different from the best quality assessment results. The VDCNN reconstruction method presents the best result for both iPhone and Nokia images showing the robustness of this method for different databases since it was the best approach for the CASIA Interval database as well. Some of the results (specially the VDCNN and PCA methods) surpass the recognition results from the original database which means that blurring the texture can be beneficial to the recognition. The PCA approach that presents good quality results does not present the best result in the recognition experiments as well as the SRGAN approach, showing that a good photo-realism does not reflect directly in a good recognition approach. Using the SIFT comparator that is based mostly on the edges and small shapes of the images, the PCA approach presents good results followed by the SRGAN approach that presents, as mentioned before, a good photo-realism in the visual results (Fig. 9 In the case of the SIFT features, the results from the PCA approach is much superior to the other methods mainly because of two facts: the PCA approach uses the same database for training the images which can generate artefacts that can differentiate the patterns in the recognition, the other fact is that PCA uses overlapping patches producing less blurred images which can be observed in Fig. 5 that can help the SIFT operator to find the key points.

We also show the density distribution curves for the VSIRIS database using the (VDCNN) and SRGAN trained with the DTD database for the reconstruction and using the QSW features in Fig. 10. It can be seen in Figs. 10*a* and *d* that even for the original databases the curves are overlapped which means that the noise contained in the original images interfere in the segmentation and in the feature extraction.

Using mobile images, the distribution of genuine scores is shifted towards the impostor distribution for all the CNN approaches as resolution decreases. This means that the 'similarity' between the reconstructed HR images and original HR images is reduced as the size of the input LR images is reduced. In other words, when the pixels of the interpolated images used as input are convolved specially for the VDCNN, this CNN does not fully resemble the information found in the original HR image (at least measured by the features employed).

## 8 Conclusion

DLSR through CNNs has been extensively explored to provide photo-realistic images from LR images (mainly from natural scenes) obtaining impressive results. Meanwhile, more relaxed acquisition circumstances, such as iris recognition through mobile phones or iris recognition through surveillance videos are boosting

76*IET Biom.*, 2019, Vol. 8 Iss. 1, pp. 69-78
This is an open access article published by the IET under the Creative Commons Attribution License
(http://creativecommons.org/licenses/by/3.0/)

the need for SR methods to improve the iris recognition process. In this study, we explore different CNN architectures that are proven to be effective in reconstructing natural images for iris SR.

In the first part of the experiments, we choose a target database (CASIA Interval) and train the most basic CNN (SRCNN) to test the texture transfer learning between different databases from different natures: natural scenes, texture images or iris images. We also perform the training using the same database divided into training (registration images) and testing (validation images) to see if training the CNN with images from the same user can help in the SR. From these experiments, we conclude that texture databases are more suitable to train the CNN for iris recognition as well as the use of the same database can also contribute to better results.

In the second part of the experiments, we chose two different databases (CASIA Interval and DTD databases) to explore the use of three different SRCNNs: SRCNN, VDCNN, and SRGAN. It can be seen that CNNs can produce more photo-realistic images with better quality than the traditional approaches. In specific, the VDCNN presents the best results in terms of quality, however, this does not reflect in terms of recognition rate. The visual analysis helps to understand this disparity in the results: where there is much photo-realism (e.g. in the case of SRGAN), there are also too many artefacts that can lead to poor results in the feature extraction. The balance between both photo-realism and smoothing images (as in the case of SRGAN for factor 8) is the perfect match for a good result.

In the last part of the experiments, we test the use of DLSR for very LR images from mobile devices trying to simulate a real-world situation. Also, in this experiment, there is a dichotomy between the quality assessment and the recognition results showing that a good photo-realism does actually not lead to a better recognition performance especially for very LR images. In the future, based on the results of this work, we intend to create and test new CNN architectures especially designed for iris SR that can provide a good balance between edge-preservation and smoothing to serve as a good pre-processing step mainly for images taken from a distance and for mobile device iris recognition systems. Also, as a future work, we will test the image re-projection and test if image enhancement on the training data can influence the reconstruction. We also intend to implement a CNN pre-trained with texture images and fine-tuned with the users' iris, which means that each user will have its own custom SRCNN that can be specifically trained for the user's iris pattern which hopefully can improve the quality of the super-resolved image even more.

## 9 Acknowledgment

This research was partially supported by CNPq-Brazil for Eduardo Ribeiro under grant no. 00736/2014-0. This project has received funding from the European Union's Horizon 2020 research and innovation program under grant agreement No 700259.